\documentclass[10pt,twocolumn,letterpaper]{article}

\usepackage{cvpr}              

\usepackage{graphicx}
\usepackage{amsmath}
\usepackage{amssymb}
\usepackage{booktabs}
\usepackage{commath} 
\usepackage[linesnumbered,ruled,vlined]{algorithm2e} 
\usepackage{caption}
\usepackage{makecell}
\usepackage[table,x11names,dvipsnames,table]{xcolor}
\usepackage{tabu}

\usepackage[pagebackref,breaklinks,colorlinks]{hyperref}

\usepackage[capitalize]{cleveref}
\crefname{section}{Sec.}{Secs.}
\Crefname{section}{Section}{Sections}
\Crefname{table}{Table}{Tables}
\crefname{table}{Tab.}{Tabs.}

\def\cvprPaperID{5320} 
\def\confName{CVPR}
\def\confYear{2022}

\begin{document}

\title{QMagFace: Simple and Accurate Quality-Aware Face Recognition}

\newcommand\Mark[1]{\textsuperscript#1}

\author{Philipp Terh\"{o}rst\Mark{1}\Mark{2}, Malte Ihlefeld\Mark{3}, Marco Huber\Mark{2}\Mark{3}, Naser Damer\Mark{2}, Florian Kirchbuchner\Mark{2}\Mark{3},\\ Kiran Raja\Mark{1}, Arjan Kuijper\Mark{2}\Mark{3}\\
\Mark{1}Norwegian University of Science and Technology, Gj{\o}vik, Norway\\ 
\Mark{2}Fraunhofer Institute for Computer Graphics Research IGD, Darmstadt, Germany\\
\Mark{3}Technical University of Darmstadt, Darmstadt, Germany\\
Email: \, {philipp.terhoerst@igd.fraunhofer.de}
}

\maketitle

\begin{abstract}
In this work, we propose QMagFace, a simple and effective face recognition solution (QMagFace) that combines a quality-aware comparison score with a recognition model based on a magnitude-aware angular margin loss.
The proposed approach includes model-specific face image qualities in the comparison process to enhance the recognition performance under unconstrained circumstances.
Exploiting the linearity between the qualities and their comparison scores induced by the utilized loss, our quality-aware comparison function is simple and highly generalizable.
The experiments conducted on several face recognition databases and benchmarks demonstrate that the introduced quality-awareness leads to consistent improvements in the recognition performance.
Moreover, the proposed QMagFace approach performs especially well under challenging circumstances, such as cross-pose, cross-age, or cross-quality.
Consequently, it leads to state-of-the-art performances on several face recognition benchmarks, such as $98.50\%$ on AgeDB, $83.95\%$ on XQLFQ, and $98.74\%$ on CFP-FP.
The code for QMagFace is publicly available\footnote{\url{https://github.com/pterhoer/QMagFace}}.
\end{abstract}

\section{Introduction}
\label{sec:intro}

\begin{figure}[h]
\centering
\includegraphics[width=0.5\textwidth]{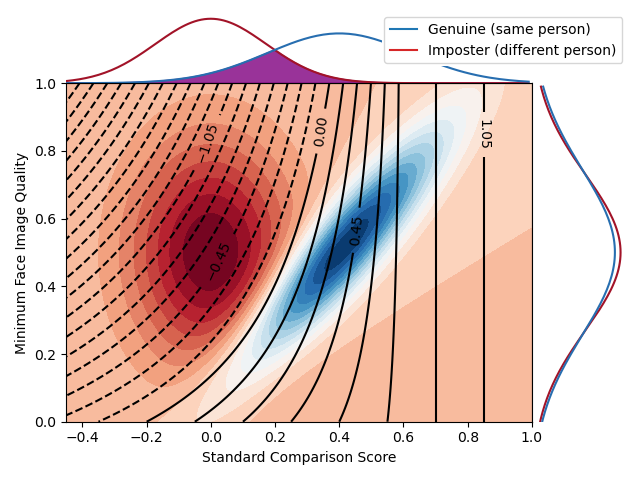}
\caption{Visualisation of the proposed quality-aware comparison score. The genuine (blue) and imposter (red) distributions are shown with respect to the comparison scores and their face image qualities. While the quality distributions (right) are very similar, the score distributions (top) are strongly overlapping. The proposed quality-aware comparison score is shown via black isolines. Dashed lines indicate negative scores and solid lines positive. The quality-awareness increases the separability of both distributions.}
\label{fig:Visualization}
\end{figure}

Face recognition systems are spreading worldwide and are increasingly involved in unconstrained environments \cite{DBLP:conf/sibgrapi/Masi0HN18}.
In these environments, these systems have to deal with large variabilities, such as challenging illuminations, poses, and expressions, that might result in incorrect  matching decisions \cite{ISO19794-5-2011}\cite{ICAO9303}.
The face image quality of a sample is defined as its utility for recognition \cite{DBLP:conf/icb/Hernandez-Ortega19}\cite{6712715}\cite{DBLP:journals/corr/Best-RowdenJ17}\cite{DBLP:conf/cvpr/TerhorstKDKK20} and measures the impact of these variabilities on the face recognition performance.
Previous works either do not employ face image quality information during comparison \cite{DBLP:conf/cvpr/LiuWYLRS17}\cite{DBLP:conf/cvpr/WangWZJGZL018}\cite{DBLP:conf/cvpr/DengGXZ19}\cite{DBLP:conf/cvpr/HuangWT0SLLH20} or include quality estimates in comparison process that are not inherently suitable for such a task \cite{DBLP:conf/iccv/ShiJ19}\cite{DBLP:conf/cvpr/LiuT21}.
While the first case results in a loss of valuable information for the comparison, in the second case, the limiting factor lies in the utilized quality estimates.

In this work, we propose QMagFace, a solution that combines a quality-aware comparison function with a face recognition model trained with a magnitude-aware angular margin (MagFace) loss.
Incorporating model-specific face image qualities in the comparison process aims at enabling an improved face recognition performance even under challenging circumstances.
Exploiting the linear relationship between the qualities and their comparison scores that is induced by the MagFace loss, our quality-aware comparison function is simple but effective.

In Figure \ref{fig:Visualization}, the effect of the proposed quality-aware scoring function is visualized.
Using the standard comparison score, the genuine (same person) and imposter (different person) distributions are strongly overlapping (see top plot).
Even if their respective quality distributions are very similar (right plot), combining both information with the proposed quality-aware scoring function increases the separability leading to more reliable comparison scores for matching.
Especially for lower comparisons and quality scores, the proposed approach adapts the scores more strongly to increase the accuracy.
For higher qualities and comparison scores, the proposed solution does not alter the score since high comparison scores imply face pairs of high quality.
Consequently, QMagFace is especially effective when dealing with challenging circumstances such as cross-pose or cross-age.

The experiments were conducted on four face recognition databases and six benchmarks.
The results demonstrate
(a) a constant improvement in the face recognition performance compared to standard comparison scores over a wide range of false match rates,
(b) the suitability of the used linear function for quality weighting,
and (c) a high generalizability of the proposed approach despite changes in backbone architecture, training databases, and evaluation benchmarks.
Moreover, the QMagFace consistently reaches high performances in video-based recognition tasks and achieves state-of-the-art results on three of the four image-based face recognition benchmarks.
Especially under challenging circumstances, such as cross-pose, cross-age, or cross-quality, QMagFace achieved high performances.

%

\section{Related Works}
\label{sec:RelatedWorks}

\subsection{Face Image Quality Assessment}

Driven by the international standards, such as ISO/IEC 19794-5 \cite{ISO19794-5-2011} and ICAO 9303 \cite{ICAO9303}, the first generation of face image quality assessment (FIQA) approaches are built on human perceptive image quality factors \cite{10.1007/978-3-540-74549-5_26}\cite{6197711}\cite{5424029}\cite{6712715}\cite{6985846}\cite{6460821} \cite{6996248}.
The second generation of FIQA approaches \cite{5981881}\cite{5981784}\cite{6877651}\cite{7351562}\cite{DBLP:journals/corr/Best-RowdenJ17}\cite{DBLP:conf/icb/Hernandez-Ortega19}\cite{Ou_2021_CVPR}\cite{DBLP:journals/corr/abs-2009-00603}  consists of supervised learning algorithms based on human or artificially constructed quality labels.
However, humans may not know the best characteristics for face recognition systems and artificially labelled quality values, derived from comparison scores, rely on error-prone labelling mechanisms.

The third generation of FIQA approaches \cite{DBLP:conf/cvpr/TerhorstKDKK20} completely avoids the use of quality labels by utilizing the face recognition networks themselves.
In 2020, Terh\"{o}rst et al. \cite{DBLP:conf/cvpr/TerhorstKDKK20} proposed stochastic embedding robustness for FIQA (SER-FIQ).
This concept measures the robustness of a face representation against dropout variations and uses this measure to determine the quality of a face.
It avoids the need for training and takes into account the decision patterns of the deployed face recognition model.
In 2021, Meng et al. \cite{DBLP:conf/cvpr/MengZH021} proposed a class of loss functions that include magnitude-aware angular margins encoding the quality into the face representation.
Training with this loss results in a face recognition model that produces embeddings whose magnitudes can measure the FIQ of their faces.

While the first two generations of FIQA methods aim to assess the utility of an image for face recognition in general, the third generation aims at determining model-specific quality values.
Therefore, we assert that these methods, SER-FIQ \cite{DBLP:conf/cvpr/TerhorstKDKK20} and MagFace \cite{DBLP:conf/cvpr/MengZH021}, have the highest potential for improving the recognition performance.
%
SER-FIQ can be applied to arbitrary face recognition networks but produce a complex relation between FIQ and comparison scores.
On the other hand, MagFace produces a linear relationship between the qualities and the comparison scores, as we will show in Section \ref{sec:Robustness Analysis}, and thus, it is more suitable for a generalisable enhancement of the recognition performance.


\subsection{Face Recognition}

In recent years, face recognition is strongly driven by advances in deep representation learning.
Early works rely on metric-learning based losses \cite{DBLP:conf/cvpr/MengZH021}, such as contrastive loss \cite{DBLP:conf/cvpr/ChopraHL05}, triplet loss \cite{DBLP:conf/cvpr/SchroffKP15}, or penalty angular-margin losses \cite{DBLP:conf/iccv/WangZWLL17}\cite{DBLP:conf/cvpr/DengGXZ19}.
However, due to the combinatorial explosion in the number of face triplets needed for training, the research focus shifted to classification-based approaches.
These approaches are typically based on softmax and aim at classifying on a closed-set of identities during training and utilize the previous layer as a feature extractor for unseen faces.
Combining the softmax activation with cross-entropy loss, most face recognition losses $\mathcal{L}$ are of the form
\begin{align}
\mathcal{L} &= - \dfrac{1}{N} \sum_{i=1}^N \log \left( \mathcal{L}_i \right) \qquad \text{with} \label{eq:FR_SoftMax_Loss} \\
\mathcal{L}_i &= \dfrac{e^{r \, \cos(m_1 \theta_{y_i}+m_2)-m_3}}{e^{r \, \cos(m_1 \theta_{y_i}+m_2)-m_3} + \sum_{j=1, j\neq y_i}^N e^{r \, \cos(\theta_{y_i})}}.
\end{align}
Here, the training set contains $N$ samples and $\Theta_{y_i}$ refers to the angle between between last layer weight-vector and the normalized feature vector $x_i$ (with $\norm{x_i}_2=r$).
For training without margins $m_1 = 1, m_2 = m_3 = 0$, this refers to a simplified softmax loss.
In SphereFace \cite{DBLP:conf/cvpr/LiuWYLRS17}, a multiplicative angular margin is deployed with $m_1 = \alpha$ and $m_2=m_3=0$.
For keeping the cosine margin penalty $m_2=\alpha$ ($m_1=1$ and $m_3=0$), this refers to CosFace \cite{DBLP:conf/cvpr/WangWZJGZL018} and for penalizing an angular margin $m_2=\alpha$ ($m_1=1$ and $m_3=0$), this refers to the ArcFace \cite{DBLP:conf/cvpr/DengGXZ19} loss.
However, these losses select a fixed margin $\alpha$ assuming that the samples are equally distributed in the embedding space around the class centers, which is not true when dealing with largely intra-class variations.
To solve this problem, solutions based on variable margins are proposed.
In \cite{DBLP:journals/corr/abs-2109-09416}, Boutros et al. proposed ElasticFace in which random margins are drawn from a Gaussian distribution $\mathcal{N}$ in each training iteration.
This allows ElasticFace to extract and retract the margins individually for each class (e.g. $m_1 = 1$ and $m_2 \thicksim N$).
Similarly, CurricularFace \cite{DBLP:conf/cvpr/HuangWT0SLLH20} addresses easy samples in the early training stage and hard ones in the later stage adapting $m_2$ while keeping $m_1=1$ and $m_3=0$.
In MagFace \cite{DBLP:conf/cvpr/MengZH021}, a magnitude-aware angular margin $m(r)$ (with $m_1=1$ and $m_3=0$) with a regularization $g(r)$ is proposed that aims at including the utility (face image quality) of a sample in the margin.
While the regularization $g(r)$ rewards samples with large magnitudes $r$, $m(r)$ is a simple linear function that aims at concentrating high-quality samples in a small region around the class centers.
This results in more suitable margins that are based on the utility of the samples and are encoded in the magnitude of the embeddings.

\subsection{Quality-Aware Face Recognition}

The qualities of face images are often used in video-to-video recognition tasks where a set of images from one person is matched to a set of images from another \cite{DBLP:conf/cvpr/LiuYO17}\cite{Ou_2021_CVPR}\cite{DBLP:journals/corr/abs-2009-00603}\cite{DBLP:journals/corr/abs-1804-01159}\cite{DBLP:conf/cvpr/YangRZCWLH17}.
There, the quality of each image is used for a weighted aggregation of information.
For recognition tasks based on single images, only a few works included the face image quality to enhance the face recognition performance.
In EQFace \cite{DBLP:conf/cvpr/LiuT21}, Liu et al. attached a quality-prediction network on a face recognition model to include the qualities in the training process.
However, this method is limited by computationally-expensive training that is not end-to-end.

In \cite{DBLP:conf/iccv/ShiJ19}, Shi and Jain proposed probabilistic face embeddings (PFE).
Representing face images as Gaussian distributions in the embedding space, the variance of each feature is interpreted as its uncertainty and thus, as its quality.
For comparison, they make use of a mutual-likelihood score to include the quality in the comparison score.
However, performance is determined by the quality estimate that is limited by (a) the used uncertainty estimation module which is trained separately from the face recognition network and (b) the assumption that each feature can be independently represented as a Gaussian Process.

In contrast to previous works, we make use of model-specific quality estimates that were linearly included in an end-to-end fashion.
Consequently, this allows our proposed solution to work simplistically and more efficiently.

\section{Methodology}
\label{sec:Methodology}

The main contribution of this work, QMagFace, combines a quality-aware comparison function with a face recognition model trained with MagFace loss \cite{DBLP:conf/cvpr/MengZH021}.
Including the model-specific face image quality in the comparison process aims to consistently improve the face recognition performance, especially under challenging conditions such as cross-pose, cross-age, or cross-quality.
Moreover, the proposed quality-aware scoring function can be robustly trained on any face recognition network based on the MagFace loss as it will be shown in Section \ref{sec:Robustness Analysis}.

\subsection{Quality-Aware Comparison Scoring}

In face biometrics, a comparison score reflects the identity similarity of two face images.
This reflection of similarity is more accurate when the compared samples are of high quality \cite{DBLP:conf/icb/Hernandez-Ortega19}.
The biometric quality of a face image is defined as its utility for recognition \cite{DBLP:conf/icb/Hernandez-Ortega19}\cite{6712715}\cite{DBLP:journals/corr/Best-RowdenJ17}\cite{DBLP:conf/cvpr/TerhorstKDKK20}.
In \cite{DBLP:conf/icb/TerhorstKDKK20}, it was shown that model-specific quality assessment reflects challenging situations for the face recognition model, such as comparisons with strong variations in pose or age.
Similar to SER-FIQ \cite{DBLP:conf/cvpr/TerhorstKDKK20}, the qualities of MagFace networks \cite{DBLP:conf/cvpr/MengZH021} utilize the deployed face recognition system and thus, strongly reflect the decision patterns and model biases. 
Consequently, we propose a simple, but effective, comparison function that includes these model-specific quality values to enhance the accuracy and robustness of the face recognition system.

Given a face recognition model $\mathcal{M}$ trained with MagFace loss and two face images $I_1$ and $I_2$, their embeddings are given by $e_1=\mathcal{M}(I_1)$ and $e_2=\mathcal{M}(I_2)$ together with the corresponding face image qualities $q_1 = \norm{e_1}_2$ and $q_2 = \norm{e_2}_2$ encoded through the vector length.
The standard comparison score $s = \cos(e_1, e_2)$ is defined over cosine similarity of these templates and thus, represents the angular similarity between both templates.
However, comparisons with low-quality images affect the comparison scores and thus, needs to be adjusted.
The proposed  quality-aware comparison score $\hat{s}$ is defined as
\begin{align}
\hat{s}(s, q_1, q_2) &= \omega(s) * \min\{q_1,q_2\} + s \label{eq:QualityAwareScore} 
\end{align}
with a quality-weighting function
\begin{align}
\omega(s) = \min \left\lbrace 0, \, \beta * s -\alpha   \right\rbrace. \label{eq:QualityWeightingFunction}
\end{align}
This comparison function consists of only two trainable parameters ($\alpha$ and $\beta$) and thus, can be robustly trained.

Since the biometric sample quality is included linearly in the MagFace loss (through $m(r)$), we assume a linear relationship between the face image qualities and their comparison scores.
In Section \ref{sec:Robustness Analysis}, we will demonstrate the suitability of a linear function for the quality weighting.
We further assume that the score-adjustment is dependent on the lowest quality of the comparison and that only comparisons with at least one low-quality sample needs to be adjusted.
Please note that the qualities $q_1$ and $q_2$ are not easily exchangeable with other FIQ methods since these need to be model-specific with regard to the FR model and require a linear relationship between the quality estimates and its comparison scores.

%
%

A high similarity score $s$ can only be achieved through the comparison of two high-quality samples.
In this case, the similarity is well reflected in the comparison score $s$ and thus, no quality-based score adjustments ($\omega = 0$) is needed.
A lower comparison score might result from the degradation of a pair with at least one low-quality sample.
In this case, the similarity is altered by the sample quality and our proposed function adjusts the score based on the quality.
Consequently, if two comparisons result in similar comparison scores $s_1 \approx s_2$  (with $\omega(s)<o$) but have different minimum qualities $q^{min}_1 \gg q^{min}_2$, the score with the higher quality undergoes a stronger adjustment and thus, results in a lower quality-aware comparison score ($\hat{s}_1 < \hat{s}_2$).

\subsection{Training the Quality-Weighting Function}

For training, the comparison scores of the training set $\mathcal{S}=\mathcal{S}_G + \mathcal{S}_I$ are separated into genuine and imposter comparisons with the corresponding minimal qualities $\mathcal{Q}^{min}$ of the sample pairs.
The training process is divided into three steps and aims at learning the quality weighting function $\omega(s)$ as shown in Algorithm \ref{algo:QualityWeightingFunction}.

In the first step, we define the optimal quality weight $w_{opt}(t)$ for a given threshold $t$.
This is given through 
\begin{align}
w_{opt}(t) = \operatorname*{argmin}_{\omega} \dfrac{1}{\abs{\mathcal{S}_{G}}} \sum_{s \in \mathcal{S}_{G}} \Theta (t - \bar{s}(\omega, s)) \label{eq:OptimalQualityWeight}
\end{align}
where $\Theta(\cdot)$ describes the Heaviside function and 
\begin{align}
\bar{s}(\omega, s) = \sigma(\omega*Q^{min}(s) + s).
\end{align}
is a quality-aware scoring function given by $\omega$ and $s$ that is scaled to the range of $[0,1]$, similar to the range of comparison scores, with a sigmoid function.
This optimization aims at minimizing the FNMR at a threshold $t$ through including face image quality information.

In the second step, a relevant threshold range $\mathcal{T}$ needs to be defined that represents the target FMR range.
In this work, we choose the range from \textit{FMR}$^{max}=10^{-2}$ to \textit{FMR}$^{min}=10^{-5}$ to cover a wide variety of potential applications and due to the amount of training data available (in the order of $10^{5}$ images).
The relevant threshold range 
$
\mathcal{T}=[ t(\textit{FMR}^{max}), t(\textit{FMR}^{min})  ]
$
is determined by finding the threshold $t$ that corresponds to the required FMR on the quality-aware scores on the training data.
This can be determined by
\begin{align*}
t(\textit{FMR}) = \operatorname*{argmin}_t \abs{ \textit{FMR} - \dfrac{1}{\abs{\mathcal{S}_{I}}} \sum_{s \in \mathcal{S}_{I}} \Theta(\bar{s}(\omega_{opt}(t), s)-t)}.
\end{align*}

The third step aims to learn the quality weighting function $\omega(s)$.
Since the quality is included linearly in the MagFace loss, a linear relationship between the importance of the quality $q(s)$ and its comparison score $s$ is assumed.
Therefore, we model the quality weights through a simple linear function 
$\omega(s) =   \beta * s - \alpha$.
The parameters $\alpha$ and $\beta$ can be learned by solving the following optimization 
\begin{align}
\min_{\alpha, \beta} \sum_{t \in \mathcal{T}} \left( \omega_{opt}(t) + \alpha - \beta * t \right)^2,
\end{align}
resulting in the optimal parameters
\begin{align}
\hat{\beta} &= \dfrac{\sum_{t \in \mathcal{T}} (t - \bar{t}) (\omega_{opt}(t) - \bar{\omega}_{opt}) }{\sum_{t \in \mathcal{T}} (t - \bar{t})^2} \\
\hat{\alpha} &=  \hat{\beta} * \bar{t} - \bar{\omega}_{opt} \quad \text{with} \\
\bar{\omega}_{opt} &= \dfrac{1}{\abs{\mathcal{T}}} \sum_{t \in \mathcal{T}} \omega_{opt}(t) \quad \text{and} \quad \bar{t} = \dfrac{1}{\abs{\mathcal{T}}} \sum_{t \in \mathcal{T}} t.
\end{align}


\begin{algorithm}
\SetKwInput{KwInput}{Input}                
\SetKwInput{KwOutput}{Output}              

\DontPrintSemicolon
  
  \KwInput{$\mathcal{S}_G, \mathcal{S}_I$, $Q^{min}$, \textit{FMR}$^{min}$, \textit{FMR}$^{max}$ }
  \KwOutput{$\omega(s)$}



\tcc{Determine optimal weights}
\For{$t \in \mathcal{T} \in [ t(\textit{FMR}^{max}), t(\textit{FMR}^{min})]$}    
        { 
        	$w_{opt}(t) \gets \operatorname*{argmin}_{\omega} \frac{1}{\abs{\mathcal{S}_{G}}} \sum_{s \in \mathcal{S}_{G}} \Theta (t - (\omega*Q^{min}(s) + s)) $
        }
\tcc{Learns quality weighting function parameters}
$\bar{\omega}_{opt} \gets \dfrac{1}{\abs{\mathcal{T}}} \sum_{t \in \mathcal{T}} \omega_{opt}(t)$\\
$\bar{t} \gets \dfrac{1}{\abs{\mathcal{T}}} \sum_{t \in \mathcal{T}} t$\\
$\hat{\beta} = \dfrac{\sum_{t \in \mathcal{T}} (t - \bar{t}) (\omega_{opt}(t) - \bar{\omega}_{opt}) }{\sum_{t \in \mathcal{T}} (t - \bar{t})^2}$ \\
$\hat{\alpha} =  \hat{\beta} * \bar{t} - \bar{\omega}_{opt}$ \\
\tcc{Define the quality function}
$\omega(s) = \min \left\lbrace 0, \, \hat{\beta} * s - \hat{\alpha}  \right\rbrace$\\
\Return $\omega(s)$
\caption{Quality-weighting function}
\label{algo:QualityWeightingFunction}
\end{algorithm}

\section{Experimental Setup}
\label{sec:ExperimentalSetup}

\subsection{Databases and Benchmarks}

To compare the performance of the proposed QMagFace approach with ten recent state-of-the-art approaches six face recognition benchmarks are used, LFW \cite{LFWTech}, AgeDB-30 \cite{DBLP:conf/cvpr/MoschoglouPSDKZ17}, CFP-FP \cite{DBLP:conf/wacv/SenguptaCCPCJ16}, XQLFW \cite{DBLP:journals/corr/abs-2108-10290}, IJB-B \cite{DBLP:conf/cvpr/WhitelamTBMAMKJ17}, and IJB-C\cite{DBLP:conf/icb/MazeADKMO0NACG18}.

LFW \cite{LFWTech} is a face verification benchmark containing 13k images of over 5k identities. 
In the benchmark experiments, we followed the standard protocol \cite{LFWTech} using the 6k predefined comparison pairs.
Moreover, we conducted the experiments on three more challenging benchmarks representing the issues of cross-age (AgeDB \cite{DBLP:conf/cvpr/MoschoglouPSDKZ17}), cross-pose (CFP-FP \cite{DBLP:conf/wacv/SenguptaCCPCJ16}), and cross-quality (XQLFW \cite{DBLP:journals/corr/abs-2108-10290}).
AgeDB \cite{DBLP:conf/cvpr/MoschoglouPSDKZ17} is unconstrained face recognition benchmark for age-invariant face verification.
It contains over 16k images of over 5k identities.
In the experiments, we follow the protocol of AgeDB-30 since it is the most reported and challenging one for AgeDB consisting of age gaps of over 30 years.
CFP-FP \cite{DBLP:conf/wacv/SenguptaCCPCJ16} is a face recognition benchmark that addresses the issue of comparing frontal to profile face images.
In our experiments, we followed the evaluation protocol of \cite{DBLP:conf/wacv/SenguptaCCPCJ16} containing 3500 genuine pairs and 3500 imposter pairs.
XQLFW \cite{DBLP:journals/corr/abs-2108-10290} is a benchmark that addresses the problem of cross-quality comparisons in face recognition.
The protocol defines 6k face image pairs based on the LFW database.
However, for each pair, one face image is of much lower quality than the other face. 
The IARPA Janus Benchmark-B (IJB-B) \cite{DBLP:conf/cvpr/WhitelamTBMAMKJ17} contains around 21k images and 55k frames from over 7k videos of 1,845 identities. 
In the experiment, we follow the standard evaluation protocol using around 10k genuine and 8M impostor comparisons.
The IARPA Janus Benchmark–C (IJB-C) \cite{DBLP:conf/icb/MazeADKMO0NACG18} extends on
the IJB-B by adding more identities.
In total, it consists of 31k images with over 117k frames of over 11k videos from 3531 identities.
The verification protocol considers over 19k genuine and 16M imposter comparisons.
In contrast to IJB-B, IJB-C focuses more on occlusion and diversity of subject occupation to improve the representation of the global population.


Besides reporting the verification accuracy based on benchmarks, we make use of four face recognition datasets to cover a much wider range of possible decision thresholds and thus, to cover more potential applications.
Morph \cite{DBLP:conf/fgr/RicanekT06} consists of 55k face images from over 13k subjects.
The images are frontal and of high quality.
LFW \cite{LFWTech} contains 13k images of over 5k identities and the ColorFeret database \cite{DBLP:journals/pami/PhillipsMRR00} consists of 14k high-resolution face images from over 1k different individuals.
The data possess a variety of poses (from frontal to profile) and facial expressions under well-controlled conditions.
The Adience dataset \cite{Eidinger:2014:AGE:2771306.2772049} consists of 26k images from over 2k different subjects.
The images of the Adience dataset possess a wide range in terms of image quality.
In the supplementary material, we included a more detailed discussion, such as on the licenses.

\subsection{Evaluation Metrics}

Following the international standard for biometric verification evaluation \cite{ISO_Metrik}, we report the face verification error in terms of false non-match rate (FNMR) at fixed false match rate (FMR).
Moreover, we report the equal error rate (EER) and the area under curve (AUC) of the receiver operating characteristic (ROC) curve.
The EER equals the FMR at the threshold where FMR = FNMR and is well known as a single-value indicator of the verification performance.
In our experiments, we report the face verification performance over a wide range of FMRs to cover a variety of potential applications.
On the benchmarks, we follow the mentioned protocols and report the verification accuracy to be comparable with previous works.

\subsection{Face Recognition Models}

In the experiments, the proposed QMagFace approach is built on three pre-trained models\footnote{\url{https://github.com/IrvingMeng/MagFace} (Apache License 2.0)} based on MagFace loss released by the authors \cite{DBLP:conf/cvpr/MengZH021}.
These were trained on the MS1MV2 database \cite{DBLP:conf/eccv/GuoZHHG16} and are based on iResNet-18, iResNet-50, and iResNet-100 backbones \cite{DBLP:conf/icpr/DutaL0020}.
In the following, we use the name of the loss function and the model trained with it interchangeably to keep this work easily comprehensible.

\begin{table}
\renewcommand{\arraystretch}{1.053}
\centering
\caption{Learned parameters}
\label{tab:Parameters} 
\begin{tabular}{lcc}
\Xhline{2\arrayrulewidth}
 & \multicolumn{2}{c}{Learned parameters}                       \\
 \cmidrule(rl){2-3}
Model                 & $\alpha$ & $\beta$ \\
\hline
QMagFace-18    & 0.092861 & 0.135311 \\
QMagFace-50    & 0.065984 & 0.103799 \\
QMagFace-100   & 0.077428 & 0.125926 \\
\Xhline{2\arrayrulewidth}
\end{tabular}
\end{table}

The parameters $\alpha$ and $\beta$ needed for the proposed QMagFace approach are trained on the Adience dataset \cite{Eidinger:2014:AGE:2771306.2772049} due to the large quality variance in its samples and to create a generalizable approach by using the estimated parameters on this dataset.
However, the training process is robust and thus, the choice of the training database only affect the performance minimally as we will show in Section \ref{sec:Robustness Analysis}.
Due to the simplicity of the proposed approach, the learned parameters are shown in Table \ref{tab:Parameters}.
To extract a face embedding from a given face image, the image is aligned, scaled, and cropped as described in \cite{DBLP:conf/cvpr/MengZH021}.
Then, the preprocessed image is passed to the face recognition models to extract the feature embedding.

%
%

\subsection{Investigations}
The proposed approach is analysed in three steps.
First, we report the performance of QMagFace on six face recognition benchmarks against ten recent state-of-the-art methods in image- and video-based recognition tasks to provide a comprehensive comparison with state-of-the-art.
Second, we investigate the face recognition performance of QMagFace over a wide FMR range to show its suitability for a wide variety of applications and to demonstrate that the quality-aware comparison score constantly enhances the recognition performance.
Third, we analyse the optimal quality weight over a wide threshold range to demonstrate the robustness of the training process and the generalizability of the proposed approach.

\section{Results}
\label{sec:Results}

\subsection{Performance on Single-Image Benchmarks}
\label{sec:Benchmarks}

To demonstrate that the proposed QMagFace approach achieves state-of-the-art performance in image-to-image face recognition tasks, the proposed method is compared against ten recent face recognition models on four benchmarks.
For PFE\footnote{\url{https://github.com/seasonSH/Probabilistic-Face-Embeddings} (MIT License)} \cite{DBLP:conf/iccv/ShiJ19}, ArcFace\footnote{\url{https://github.com/deepinsight/insightface} (MIT License)} \cite{DBLP:conf/cvpr/DengGXZ19}, and the MagFace variants \cite{DBLP:conf/cvpr/MengZH021}, we used the implementations released by the authors.
The remaining benchmark results are taken from \cite{DBLP:journals/corr/abs-2109-09416}\cite{DBLP:conf/cvpr/LiuT21}.
In Table \ref{tab:Benchmarks}, the face recognition performances of these are shown.
On the LFW benchmark, the proposed QMagFace approach based on the iResNet-100 backbone achieved a performance of 99.83\%, which is close to the state-of-the-art performance of 99.85\%.
On more challenging and less-saturated benchmarks, the proposed approach achieves state-of-the-art performance.
This includes 98.50\% on cross-age face recognition (AgeDB), 98.74\% on cross-pose face recognition (CFP-FP), and 83.95\% on cross-quality face recognition (XQLFW).
Since the FIQ captures these challenging conditions and the quality values represent the utility of the images for our specific network, the proposed quality-aware comparison score can specifically address the circumstance and their effect on the network.
Consequently, it performs highly accurate in the cross-age, cross-pose, and cross-quality scenarios and achieves state-of-the-art performances.

\begin{table}[t]
\renewcommand{\arraystretch}{1.1}
\setlength{\tabcolsep}{2.8pt}
\centering
\caption{Image-to-image face recognition performance on four benchmarks reported in terms of benchmark accuracy ($\%$). The highest performance is marked bold. The proposed  approach, QMagFace-100, achieves state-of-the-art face recognition performance, especially in cross-age (AgeDB), cross-pose (CFP-FP), and cross-quality (XQLFW) scenarios.}
\label{tab:Benchmarks} 
\begin{tabular}{lcccc}
\Xhline{2\arrayrulewidth}
 & \multicolumn{4}{c}{Benchmark}                       \\
 \cmidrule(rl){2-5}
Model                 & AgeDB & CFP-FP & LFW   & XQLFW \\
\hline
SphereFace \cite{DBLP:conf/cvpr/LiuWYLRS17}            & 98.17 & 86.84  & 99.67 &  -     \\
CosFace \cite{DBLP:conf/cvpr/WangWZJGZL018} & 98.17 & 98.26 & 99.78 &- \\
PFE \cite{DBLP:conf/iccv/ShiJ19} & 96.90 & 97.49 & 99.80 & 79.80 \\
ArcFace \cite{DBLP:conf/cvpr/DengGXZ19}              & 98.07 & 97.31  & 99.77 & 79.73 \\
GroupFace  \cite{DBLP:conf/cvpr/KimPRS20}           & 98.28 & 98.63  & \textbf{99.85} &   -    \\
CurricularFace \cite{DBLP:conf/cvpr/HuangWT0SLLH20}       & 98.32 & 98.37  & 99.80 &   -    \\
ElasticFace-Arc \cite{DBLP:journals/corr/abs-2109-09416}       & 98.35 & 98.67  & 99.80 & 81.87 \\
ElasticFace-Cos \cite{DBLP:journals/corr/abs-2109-09416}       & 98.27 & 98.61  & 99.82 & 83.78 \\
EQFace \cite{DBLP:conf/cvpr/LiuT21} & - & 98.34 & 99.80 & - \\
MagFace-18 \cite{DBLP:conf/cvpr/MengZH021}         & 93.37 & 93.11 & 99.22 & 69.55 \\
QMagFace-18 (ours)  & 92.98 & 94.00 & 99.30 & 68.60 \\
MagFace-50 \cite{DBLP:conf/cvpr/MengZH021}         & 97.60 & 97.33 & 99.72 & 80.60 \\
QMagFace-50 (ours)  & 97.88 & 97.74 & 99.73 & 80.63 \\
MagFace-100 \cite{DBLP:conf/cvpr/MengZH021}        & 98.18 & 98.36 & 99.73 & 83.90 \\
QMagFace-100 (ours) & \textbf{98.50} & \textbf{98.74} & 99.83 & \textbf{83.95} \\
\Xhline{2\arrayrulewidth}
\end{tabular}
\end{table}

\subsection{Performance on Video-Based Benchmarks}

In Table \ref{tab:Benchmarks_Video}, the video-based face recognition performance is analysed based on IJB-B and IJB-C.
The FNMR is investigated over a wide range of FMRs.
The performances of most state-of-the-art approaches are taken from the respective works.
Since some original works did not investigate the performance on IJB-B/C, the remaining performances are taken from \cite{DBLP:conf/cvpr/MengZH021} and \cite{DBLP:journals/corr/abs-2109-09416}.
For creating an embedding with the corresponding quality-value for a video, the unit-sized embeddings and qualities per frame are aggregated by a quality-weighted sum.
Introducing quality-awareness to the MagFace-100 model generally reduces the recognition error.
Especially for FMRs up to $10^{-3}$, QMagFace achieves state-of-the-art performance.
Despite the effectiveness of QMagFace for image-to-image face recognition tasks, the quality of a video-embedding does currently not well represent its utility for recognition.
While the quality corresponding to an embedding for a single frame has a high correlation with the true utility of this frame, the same does not apply for the (weighted-sum) aggregated embedding and thus for the quality of the video.
This needs to be addressed by future work.

\begin{table*}[t]
\renewcommand{\arraystretch}{1.1}
\setlength{\tabcolsep}{7.0pt}
\centering
\caption{Video-based face recognition performance. The performance [$\%$] is reported in terms of FNMR at different FMRs. The best is marked bold. The asterisk (*) denotes a method that is optimized for video-based recognition. In nearly all cases, the quality-awareness increases the performance of MagFace. For FMRs up to $10^{-3}$, the proposed quality-aware solution performs best despite that aggregated quality does not reflect well the combined embeddings per video.
}
\label{tab:Benchmarks_Video} 
\begin{tabular}{lcccccccc}
\Xhline{2\arrayrulewidth}
                & \multicolumn{4}{c}{IJB-B}    & \multicolumn{4}{c}{IJB-C}    \\
        \cmidrule(rl){2-5}     \cmidrule(rl){6-9}    
 FNMR at FMR of               & $10^{-2}$  & $10^{-3}$  & $10^{-4}$  & $10^{-5}$              & $10^{-2}$  & $10^{-3}$  & $10^{-4}$  & $10^{-5}$    \\
                \hline
CosFace \cite{DBLP:conf/cvpr/WangWZJGZL018} & - & - & 5.99 & 10.75 & - & - & 4.44 & 7.32 \\
PFE \cite{DBLP:conf/iccv/ShiJ19} & - & - & - & - & 2.83 & 4.51 & 6.75 & 10.36 \\
ArcFace \cite{DBLP:conf/cvpr/DengGXZ19} & 2.47 & 3.84 & 5.77 & 10.79 & 1.82 & 2.79 & 4.36 & 6.89 \\
GroupFace \cite{DBLP:conf/cvpr/KimPRS20} & - & - & 5.07 & - & - & - & 3.74 & - \\
CurricularFace \cite{DBLP:conf/cvpr/HuangWT0SLLH20} & - & - & 5.20 & - & - & - & 3.90 &  -\\
ElasticFace-Arc \cite{DBLP:journals/corr/abs-2109-09416} & - & - & 4.78 & - & - & - & 3.51 & - \\
ElasticFace-Cos \cite{DBLP:journals/corr/abs-2109-09416} & - & - & \textbf{4.70} & - & - & - & \textbf{3.43} & - \\
EQFace* (QWFA) \cite{DBLP:conf/cvpr/LiuT21} & 2.76 & 3.69 & 5.12 & \textbf{8.13} & 1.78 & 2.55 & 3.62 & \textbf{5.07} \\
EQFace (QW) \cite{DBLP:conf/cvpr/LiuT21} & 2.38 & \textbf{3.52} & 5.49 & 10.38 & 1.70 & 2.61 & 4.16 & 6.99 \\
MagFace-100 \cite{DBLP:conf/cvpr/MengZH021} & 2.57 & 3.81 & 5.50 & 9.64 & 1.74 & 2.76 & 4.03 & 5.92 \\
QMagFace-100 (ours) & \textbf{2.28} & \textbf{3.52} & 5.30 & 9.71 & \textbf{1.49} & \textbf{2.38} & 3.81 & 5.73 \\
\Xhline{2\arrayrulewidth}             
\end{tabular}
\end{table*}

\subsection{Full Performance Analysis}
\label{sec:FullPerformance}
In Table \ref{tab:Benchmarks}, it was already shown that proposed quality-aware face recognition approach leads to stable improvements in the recognition performance.
In this section, we will demonstrate these improvements for a wide FMR range.
Table \ref{tab:FullFRPerformance} shows the recognition performance of MagFace and QMagFace variants.
To cover a wide range of potential applications, the performance is analysed over a wide range of decision thresholds (ranging from FMR of $10^{-1}$ to $10^{-5}$) for three databases.
The analysis involved over 300k/160k/1.1M comparisons on the ColorFeret/LFW/Morph database.
For three backbones (iResNet-18/50/100), the performance of MagFace and the proposed QMagFace approach is compared.

For QMagFace-18, 19 out of 21 scenarios showed an improved recognition performance while for QMagFace-50, 20 out of 21 scenarios showed a performance enhancement.
The three cases with a decreased performance took place on ColorFeret, which involves many challenging frontal to profile face comparisons.
Since the recognition performance of MagFace-18 and MagFace-50 is much lower than the models based on iResNet-100, the performance in estimating the model-specific face image quality correctly is lower as well. 
Therefore, the QMagFace approaches that make use of these quality estimates become less accurate when the quality estimate is failed by a large degree.

For QMagFace-100, the performance, and thus the quality estimation, is higher.
Consequently, the proposed QMagFace-100 approach leads to strong performance improvements in all investigated cases.

\begin{table*}[t]
\renewcommand{\arraystretch}{1.1}
\setlength{\tabcolsep}{10pt}
\centering
\caption{Face recognition performance reported in terms of FNMR [$\%$] over a wide range of FMRs. The MagFace and the proposed QMagFace approach are compared for three backbones on three databases. The better values between both approaches are highlighted in bold. In general, the proposed quality-aware solutions constantly improve the performance, often by a large margin. This is especially true for QMagFace based on the iResNet-100 backbone.}
\label{tab:FullFRPerformance} 
\begin{tabular}{llccccccc}
\Xhline{2\arrayrulewidth}
 & & & \multicolumn{5}{c}{FNMR at FMR} & \\
\cmidrule(rl){4-8}
Database         & Model          & EER  & $10^{-1}$  & $10^{-2}$  & $10^{-3}$  & $10^{-4}$  & $10^{-5}$  & AUC     \\
\hline
ColorFeret & MagFace-18   & 5.312 & 4.067   & 9.193   & \textbf{22.094}  & 83.968  & \textbf{97.517}  & 98.48 \\
           & QMagFace-18  & \textbf{4.232} & \textbf{3.068}   & \textbf{6.902}   & 22.951  & \textbf{82.531}  & 97.723  & \textbf{98.65} \\
 \rowcolor{Snow2} \cellcolor{White}          & MagFace-50   & 3.635 & 2.553   & 5.056   & 7.560   & 12.393  & \textbf{22.416}  & 99.09 \\
 \rowcolor{Snow2} \cellcolor{White}          & QMagFace-50  & \textbf{2.941} & \textbf{1.464}   & \textbf{4.173}   & \textbf{6.832}   & \textbf{12.247}  & 23.426  & \textbf{99.55} \\
           & MagFace-100  & 2.629 & 1.789   & 3.297   & 4.791   & 7.523   & 16.909  & 99.24 \\
           & QMagFace-100 & \textbf{2.060} & \textbf{0.950}   & \textbf{2.616}   & \textbf{4.409}   & \textbf{7.145}   & \textbf{16.454}  & \textbf{99.67} \\
           \hline
\rowcolor{Snow2} \cellcolor{White}Morph      & MagFace-18   & 0.883 & 0.813   & 0.873   & 1.185   & 2.189   & 50.892  & 99.43 \\
\rowcolor{Snow2} \cellcolor{White}           & QMagFace-18  & \textbf{0.843} & \textbf{0.779}   & \textbf{0.834}   & \textbf{1.036}   & \textbf{1.908}   & \textbf{40.070}  & \textbf{99.53} \\
           & MagFace-50   & \textbf{0.788} & 0.784   & 0.825   & 0.832   & 0.843   & 0.894   & 99.61 \\
           & QMagFace-50  & 0.821 & \textbf{0.473}   & \textbf{0.812}   & \textbf{0.826}   & \textbf{0.835}   & \textbf{0.880}   & \textbf{99.84} \\
\rowcolor{Snow2} \cellcolor{White}           & MagFace-100  & 0.848 & 0.777   & 0.814   & 0.824   & 0.834   & 0.848   & 99.58 \\
\rowcolor{Snow2} \cellcolor{White}           & QMagFace-100 & \textbf{0.773} & \textbf{0.363}   & \textbf{0.760}   & \textbf{0.817}   & \textbf{0.829}   & \textbf{0.840}   & \textbf{99.88} \\
           \hline
LFW        & MagFace-18   & 1.057 & 0.324   & 1.096   & 3.710   & 9.613   & 20.163  & 99.86 \\
           & QMagFace-18  & \textbf{0.724} & \textbf{0.186}   & \textbf{0.607}   & \textbf{2.324}   & \textbf{7.282}   & \textbf{14.377}  & \textbf{99.93} \\
\rowcolor{Snow2} \cellcolor{White}           & MagFace-50   & 0.349 & 0.110   & 0.290   & 0.462   & 0.586   & 0.821   & 99.97 \\
\rowcolor{Snow2} \cellcolor{White}           & QMagFace-50  & \textbf{0.332} & \textbf{0.035}   & \textbf{0.172}   & \textbf{0.407}   & \textbf{0.517}   & \textbf{0.752}   & \textbf{99.99} \\
           & MagFace-100  & 0.277 & 0.159   & 0.255   & 0.297   & 0.441   & 0.621   & 99.94 \\
           & QMagFace-100 & \textbf{0.195} & \textbf{0.145}   & \textbf{0.172}   & \textbf{0.221}   & \textbf{0.331}   & \textbf{0.517}   & \textbf{99.95} \\
           \Xhline{2\arrayrulewidth}
\end{tabular}
\end{table*}

\subsection{Robustness Analysis}
\label{sec:Robustness Analysis}
Lastly, we demonstrate (a) the suitability of choosing a linear quality-weighting function and (b) the generalizability of the QMagFace solution.

\begin{figure*}[t]
\centering
\subfloat[MagFace-18\label{fig:Corr18}]{%
     \includegraphics[width=0.33\textwidth]{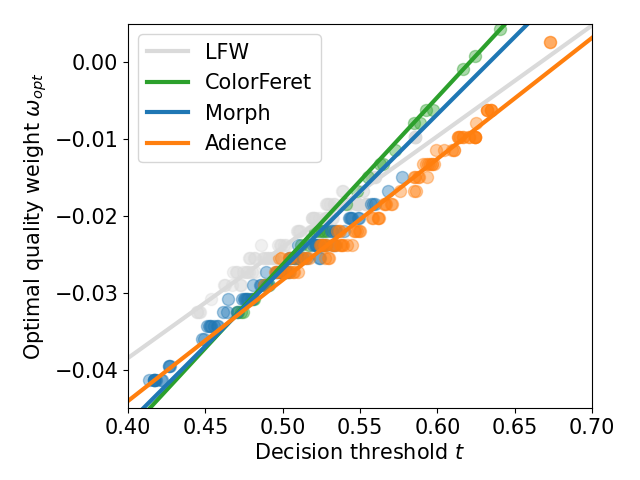}} 
\subfloat[MagFace-50 \label{fig:Corr50}]{%
     \includegraphics[width=0.33\textwidth]{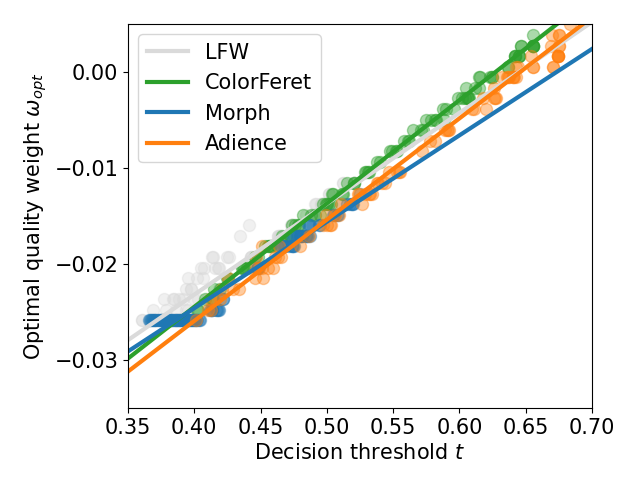}}
\subfloat[MagFace-100 \label{fig:Corr100}]{%
     \includegraphics[width=0.33\textwidth]{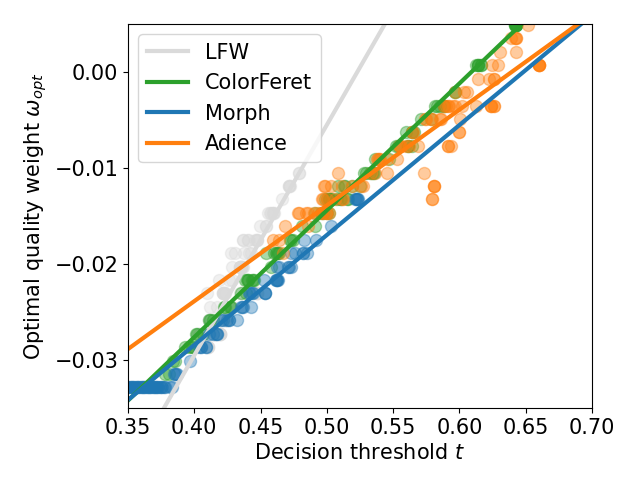}}
\caption{Optimal quality weight for different decision thresholds on four databases. The threshold range reflect FMRs from $10^{-2}$ to $10^{-5}$. Training on different databases lead to similar linear solutions for $\omega(s)$. The results demonstrate that (a) the choice of a linear function is justified and (b) that the learned models have a high generalizability since the weighting function $\omega(s)$ trained on one database is very similar to the optimal functions of the others.} 
\label{fig:CorrelationPlots}
\end{figure*}

In Figure \ref{fig:CorrelationPlots}, the correlation between the optimal quality weight and different decision thresholds are shown for different MagFace models and databases.
For each of the four databases, the optimal quality weight $\omega_{opt}$ is computed with Eq. \ref{eq:OptimalQualityWeight} for several thresholds $t\in \mathcal{T}$.
These weights show the optimal solutions for a given database that can be achieved by using the quality-aware score function $\hat{s}(s, q_1, q_2)$ from Eq. \ref{eq:QualityAwareScore}.
Moreover, a linear function is fitted through these points and shown in the same color.

For all three models (Figure \ref{fig:Corr18}, \ref{fig:Corr50}, \ref{fig:Corr100}), two observations are made.
First, the optimal quality weights $\omega_{opt}$ follow a linear function with respect to the decision threshold $t$.
This is observed for all MagFace models and on each database, proving the suitability of our linear quality weight function $\omega(s)$ from Eq. \ref{eq:QualityWeightingFunction}.
Second, for each model, the optimal quality weight functions are similar.
The only exception is the quality weight function for MagFace-100 that is optimized on LFW.
In this case, the database turns out to be too easy to train the model effectively since even for an FMR of $10^{-5}$ the decision threshold is below 0.5.
However, in all the other investigated cases, the relation between the optimal quality weights and the decision thresholds is similarly independent of the analysed database.
Utilising these databases for training, QMagFace will lead to similar matching decisions demonstrating the robustness of the QMagFace training process.
Moreover, it indicates a high generalizability since the learned function on one database is very similar to the optimal functions of the others.

\section{Limitations and Ethical Considerations}

Despite the high generalizability and the effectiveness of QMagFace for unconstrained face recognition, the approach has to deal with two limiting factors.
First, the additional quality information is most beneficial for images of lower quality and thus, the performance improvements of QMagFace decrease for very low FMRs, such as $10^{-7}$.
Second, for more effective video-based recognition, a more suitable quality aggregation is needed.
For single images, the quality of an embedding well reflects its utility for recognition.
However, this does not apply when fusing the frames of a video with the corresponding qualities.
To more efficiently exploit the quality information in video-based recognition with QMagFace, future works need to focus on more advanced quality-based fusion techniques for video frames.

While the proposed quality-awareness approach might strongly improve unconstrained face recognition for the sake of higher security or convenience, we want to point out the importance of unbiased quality estimates for fair face recognition.
The use of biased quality estimates might lead to unfair, and thus discriminatory, matching decisions depending on demographic and non-demographic factors of their users \cite{DBLP:journals/corr/abs-2103-01592}.

\section{Conclusion}

For recognition in unconstrained environments, FR systems have to deal with challenging situations, such as different illuminations, poses, and expressions.
Previous works either focused on learning margin-based approaches while not considering FIQ information or included non-inherently fit quality estimates.
In this work, we proposed QMagFace, a simple and robust quality-aware FR approach.
It integrates model-specific FIQ information in the comparison process to allow a more accurate performance under challenging situations, such as cross-pose or cross-age.
The experiments were conducted on ten FR databases and benchmarks.
The results demonstrated that including the quality-awareness consistently increases the FR performance.
Moreover, it was shown that QMagFace reaches competitive recognition results with state-of-the-art solutions. 
For challenging circumstances, such as cross-pose, cross-age, or cross-quality, QMagFace constantly beat state-of-the-art approaches.
Additional experiments indicated a high generalizability of the proposed approach demonstrated the suitability of a linear function for the quality weighting.

\section*{Acknowledgment}
This research work has been funded by the German Federal Ministry of Education and Research and the Hessen State Ministry for Higher Education, Research and the Arts within their joint support of the National Research Center for Applied Cybersecurity ATHENE.
Portions of the research in this paper use the FERET database of facial images collected under the FERET program, sponsored by the DOD Counterdrug Technology Development Program Office.
This work was carried out during the tenure of an ERCIM ’Alain Bensoussan‘ Fellowship Programme.


{\small
\bibliographystyle{ieee_fullname}
\bibliography{egbib}
}

\newpage
\clearpage

\section*{Supplementary}
In the main paper, we limited information on the databases to essential parts.
At this point, we provide additional information in the context of the used databases to enhance the understanding of this work.
More precisely, this supplementary material consists of three parts.
First, we provide information on the quality distributions of the databases to better understand the choice of the training dataset.
Second, we provide additional information on the database licenses and their creation processes.
Third, demonstrate the effect of using different training databases for QMagFace to support the reasoning concerning the generalizability of QMagFace from Section \ref{sec:Robustness Analysis} in a more direct manner.


\subsection*{Quality Distributions of MagFace}

\begin{figure*}[]
\centering
\subfloat[MagFace-18\label{fig:Quality-18}]{%
     \includegraphics[width=0.33\textwidth]{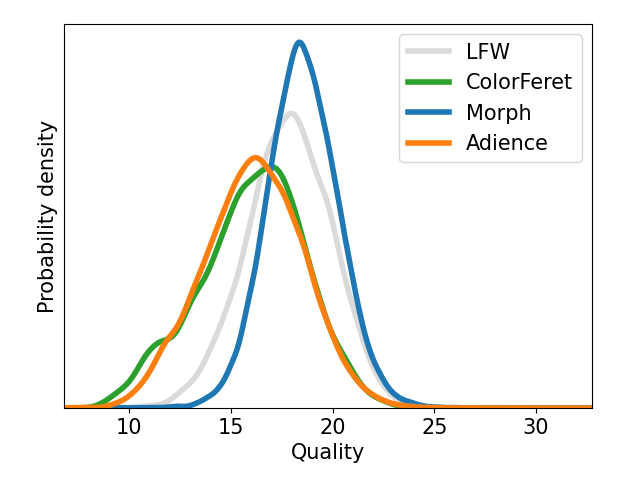}} 
\subfloat[MagFace-50 \label{fig:Quality-50}]{%
     \includegraphics[width=0.33\textwidth]{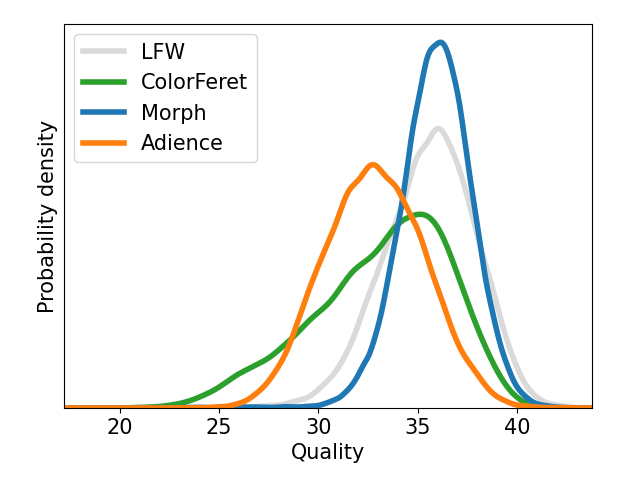}}
\subfloat[MagFace-100 \label{fig:Quality-100⌡}]{%
     \includegraphics[width=0.33\textwidth]{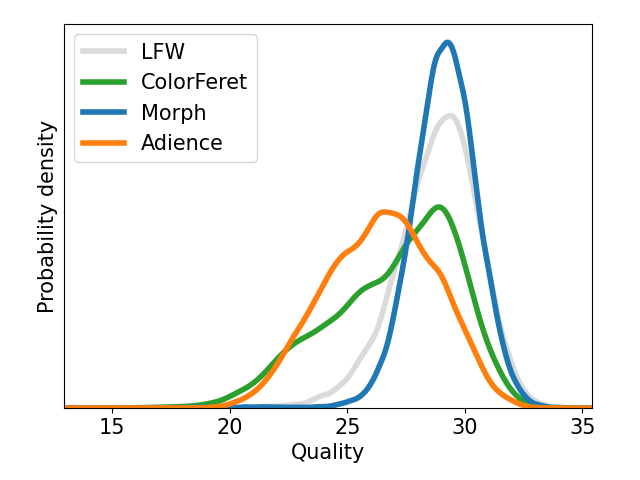}}
\caption{Quality distributions for the four FR datasets for the three MagFace backbones based on iResNet-18/50/100.} 
\label{fig:qualityDistributions}
\end{figure*}

The proposed QMagFace approach makes use of MagFace qualities and includes these in the decision process.
To get a better understanding of the quality distributions of the different used databases, Figure \ref{fig:qualityDistributions} shows these distributions for the three MagFace backbones.
For all backbones, LFW and Morph consist of the highest FIQ values and share a similar distribution due to the fact that both databases consist of mostly frontal and well-illuminated images with high image quality.
For MagFace-50 and MagFace-100, the quality distribution of ColorFeret shows the widest range of FIQ values.
ColorFeret consists of high-quality images that were taken under controlled capturing conditions.
The high variety of FIQ values origin from the head pose variations and the lowest FIQ values come from the profile face images since these prove to have a very low utility for recognition \cite{DBLP:conf/icb/TerhorstKDKK20}.
The Adience database consists of face images with a wide variety of quality-decreasing factors such as variations in image quality, occlusions, expressions, and head poses.
However, it does not contain many full profile images and thus, ColorFeret consists of images with lower FIQ values.
It should be noted that the quality estimation performance of MagFace is dependent on its FR performance.
Consequently, for MagFace-18, this leads to many wrongly assessed qualities and thus, to a lower performance of QMagFace-18.
This also explains why for MagFace-18 the quality distributions are similar while for MagFace-50 and MagFace-100 the distributions show strong differences.

\subsection*{Additional Information on the Utilized Databases}

%

After discussing the additional properties of the databases themselves, this section provides additional information about the licenses and the creation process of these databases.
Since the amount of information about the used datasets is restricted by the page limit, the paper focuses on the most important aspects to make the experiments understandable and reproducible.

LFW \cite{LFWTech} is licensed under CC-BY-4.0.
It is based on the Faces in the Wild database \cite{DBLP:conf/cvpr/BergBEMWTLF04} collected by Tamara Berg at Berkeley and consists of face captioned from news images.
More details can be found in \cite{LFWTech} or under \url{http://vis-www.cs.umass.edu/lfw/index.html}.
%

AgeDB \cite{DBLP:conf/cvpr/MoschoglouPSDKZ17} is available for non-commercial research purposes only and consists of images manually collected from the internet.
More details on the collection process can be found in \cite{DBLP:conf/cvpr/MoschoglouPSDKZ17} and the details on the license are presented in \url{https://ibug.doc.ic.ac.uk/resources/agedb/}.
%

CFP-FP \cite{DBLP:conf/wacv/SenguptaCCPCJ16} consist of manually collected images of celebrities in frontal and profile views.
More information can be found in \cite{DBLP:conf/wacv/SenguptaCCPCJ16} and \url{http://www.cfpw.io/}.
To get more information on license and consent, we reached out to the first author via mail.

XQLFW \cite{DBLP:journals/corr/abs-2108-10290} is licensed under the MIT License and is based on the modified images of the LFW dataset \cite{LFWTech} (CC-BY-4.0).
Detailed information can be found in \cite{DBLP:journals/corr/abs-2108-10290}, \url{https://martlgap.github.io/xqlfw/pages/citation.html}, and \url{https://github.com/Martlgap/xqlfw}.


The images of the IJB-B \cite{DBLP:conf/cvpr/WhitelamTBMAMKJ17} and the IJB-C \cite{DBLP:conf/icb/MazeADKMO0NACG18} databases from the National Institute for Standards and Technology (NIST) are made available under different Creative Commons
license variants.
Details on the collection process and corresponding informationcan be found in \cite{DBLP:conf/cvpr/WhitelamTBMAMKJ17} and IJB-C \cite{DBLP:conf/icb/MazeADKMO0NACG18}.
More information on the license are shown under \url{https://nigos.nist.gov/datasets/ijbc/request} and \url{https://nigos.nist.gov/facechallenges/data/IJBC/IJBC_LICENSES.TXT}.

%
%
%
%

Adience \cite{Eidinger:2014:AGE:2771306.2772049} is a database that includes a compilation of individual images which were uploaded to the internet and tagged as publicly available by the original author.
It is limited to research purposes only.
More information can be found in \cite{Eidinger:2014:AGE:2771306.2772049} and under \url{https://talhassner.github.io/home/projects/Adience/LICENSE.txt}.

%

In this work, the academic version of the Morph dataset \cite{DBLP:conf/fgr/RicanekT06} is used.
This is restricted to for research purposes only.
The legacy photographs associated with these records were taken 1962 and 1998. Digital
scans of these photographs were collected with legal considerations and IRB approval.
More information can be found in \cite{DBLP:conf/fgr/RicanekT06} and under \url{https://uncw.edu/oic/tech/morph.html}.

ColorFeret \cite{DBLP:journals/pami/PhillipsMRR00} database is restricted to face recognition research.
During the data collection, the different subjects were photographed in 15 sessions over three years under controlled conditions.
Detailed license information can be found under \url{https://www.nist.gov/system/files/documents/2019/11/25/colorferet_release_agreement.pdf}.
More details can be found in \cite{DBLP:journals/pami/PhillipsMRR00} and \url{https://www.nist.gov/itl/products-and-services/color-feret-database}.

\subsection*{Additional Experiments on Various Training Data}
After finalizing the discussion on the databases, this section aims to emphasize the high generalizability of the proposed approach against various training data.
In Section \ref{sec:Robustness Analysis}, this was already demonstrated indirectly by comparing the optimal quality-weighting functions for each database.
In this section, we show this in a more direct way by iteratively using the different FR databases for training.

In Table \ref{tab:DifferentTrainingBenchmarks}, the effect of the different training databases on the single-image FR benchmarks are shown. 
For QMagFace-18, the performance does not improve in all cases due to the limited quality estimation performance of MagFace-18 as discussed above.
In contrast, adding the quality-awareness to the MagFace-100 model improves the recognition performance independent of the utilized training data.
Moreover, it seems that the choice in the paper to use Adience for training was wrong since the performance when using the other databases for training is higher.
However, the choice for Adience was done to its large variety in quality-decreasing factors, such as occlusions, head poses, illuminations, and image qualities.
When it comes to smaller FMRs, these factors become more important and Adience might be the better choice for stable improvements in the recognition performance.

\begin{table*}[]
\renewcommand{\arraystretch}{1.1}
\setlength{\tabcolsep}{5pt}
\centering
\caption{The effect of the different training databases on the single-image FR benchmarks. The performance is reported in terms of benchmark accuracy ($\%$). For comparison, the performance of the QMagFace variants is shown against the MagFace models without quality-awareness. It turns out that choosing Adience, as done in the paper, leads to the weakest performance on these benchmarks. Consequently, the proposed  approach, QMagFace-100, achieves state-of-the-art face recognition performance independent of the training data, especially in cross-age (AgeDB), cross-pose (CFP-FP), and cross-quality (XQLFW) scenarios.}
\label{tab:DifferentTrainingBenchmarks} 
\begin{tabular}{llcccc}
\Xhline{2\arrayrulewidth}
Trained on & Model & AgeDB & CFP-FP & LFW & XQLFW \\
\hline
 & MagFace-18 & 93.37 & 93.11 & 99.22 & 69.55 \\
 & MagFace-50 & 97.60 & 97.33 & 99.72 & 80.60 \\
 & MagFace-100 & 98.18 & 98.36 & 99.73 & 83.90 \\
 \hline
Adience & QMagFace-18 & 92.98 & 94.00 & 99.30 & 68.60 \\
 & QMagFace-50 & 97.88 & 97.74 & 99.73 & 80.63 \\
 & QMagFace-100 & 98.50 & 98.74 & 99.80 & 83.97 \\
 \hline
ColorFeret & QMagFace-18 & 92.90 & 94.03 & 99.33 & 68.68 \\
 & QMagFace-50 & 97.88 & 97.80 & 99.73 & 80.63 \\
 & QMagFace-100 & 98.48 & 98.76 & 99.80 & 84.03 \\
 \hline
Morph & QMagFace-18 & 93.02 & 94.06 & 99.33 & 68.67 \\
 & QMagFace-50 & 97.95 & 97.86 & 99.73 & 80.57 \\
 & QMagFace-100 & 98.55 & 98.77 & 99.82 & 83.82 \\
 \hline
LFW & QMagFace-18 & 92.92 & 94.07 & 99.35 & 68.72 \\
 & QMagFace-50 & 97.88 & 97.86 & 99.72 & 80.58 \\
 & QMagFace-100 & 98.60 & 98.77 & 99.83 & 83.97 \\
 \Xhline{2\arrayrulewidth}
\end{tabular}
\end{table*}

In Tables \ref{tab:DifferentTrainingEvalOnAdience}, \ref{tab:DifferentTrainingEvalOnColorFeret}, \ref{tab:DifferentTrainingEvalOnLFW}, and \ref{tab:DifferentTrainingEvalOnMorph}, the effect of different training databases is analysed over a wide range of FMRs.
For low FMRs, such as $10^{-5}$, a larger variety of quality factors play important roles in enhancing the recognition performance and thus, using Adience as the training database leads to very stable performance improvements in all cases.
However, including the quality-awareness leads to strong performance improvements for most FMRs and the different training datasets generally leads to similar performances demonstrating the high generalizability of the proposed QMagFace approach.

\begin{table*}[]
\renewcommand{\arraystretch}{1.1}
\setlength{\tabcolsep}{4pt}
\centering
\caption{Evaluation on Adience based on different training datasets - The performance [$\%$] is reported in terms of FNMR at different FMRs and EER. Three MagFace variants are compared against QMagFace variants that are trained on different training sources.}
\label{tab:DifferentTrainingEvalOnAdience} 
\begin{tabular}{llrrrrrr}
\Xhline{2\arrayrulewidth}
 & & & \multicolumn{5}{c}{FNMR at FMR}  \\
\cmidrule(rl){4-8}
Training database & Model &  EER  & $10^{-1}$  & $10^{-2}$  & $10^{-3}$  & $10^{-4}$  & $10^{-5}$  \\
\hline
 & MagFace-18 & 4.505 & 2.665 & 10.639 & 28.935 & 49.982 & 74.662 \\
 & MagFace-50 & 2.432 & 1.334 & 3.463 & 8.818 & 18.396 & 44.821 \\
 & MagFace-100 & 2.291 & 1.395 & 2.926 & 5.478 & 11.211 & 30.331 \\
 \hline
ColorFeret & QMagFace-18 & 3.798 & 2.110 & 8.466 & 26.547 & 49.403 & 74.881 \\
 & QMagFace-50 & 2.371 & 1.276 & 3.310 & 8.604 & 18.386 & 44.232 \\
 & QMagFace-100 & 2.255 & 1.368 & 2.818 & 5.336 & 11.098 & 30.188 \\
 \hline
LFW & QMagFace-18 & 3.794 & 2.099 & 8.360 & 25.556 & 48.337 & 73.807 \\
 & QMagFace-50 & 2.369 & 1.286 & 3.280 & 8.479 & 19.001 & 45.802 \\
 & QMagFace-100 & 2.266 & 1.381 & 2.810 & 5.277 & 11.765 & 30.967 \\
 \hline
Morph & QMagFace-18 & 3.792 & 2.120 & 8.396 & 26.397 & 49.208 & 74.669 \\
 & QMagFace-50 & 2.368 & 1.289 & 3.282 & 8.412 & 19.405 & 45.381 \\
 & QMagFace-100 & 2.264 & 1.369 & 2.817 & 5.288 & 11.443 & 30.704 \\
 \Xhline{2\arrayrulewidth}
\end{tabular}
\end{table*}

\begin{table*}[]
\renewcommand{\arraystretch}{1.1}
\setlength{\tabcolsep}{4pt}
\centering
\caption{Evaluation on ColorFeret based on different training datasets - The performance [$\%$] is reported in terms of FNMR at different FMRs and EER. Three MagFace variants are compared against QMagFace variants that are trained on different training sources}
\label{tab:DifferentTrainingEvalOnColorFeret} 
\begin{tabular}{llrrrrrr}
\Xhline{2\arrayrulewidth}
 & & & \multicolumn{5}{c}{FNMR at FMR}  \\
\cmidrule(rl){4-8}
Training database & Model &  EER  & $10^{-1}$  & $10^{-2}$  & $10^{-3}$  & $10^{-4}$  & $10^{-5}$  \\
\hline
& MagFace-18 & 5.312 & 4.067 & 9.193 & 22.094 & 83.968 & 97.517 \\
 & MagFace-50 & 3.635 & 2.553 & 5.056 & 7.560 & 12.393 & 22.416 \\
 & MagFace-100 & 2.629 & 1.789 & 3.297 & 4.791 & 7.523 & 16.909 \\
 \hline
Adience & QMagFace-18 & 4.232 & 3.068 & 6.902 & 22.951 & 82.531 & 97.723 \\
 & QMagFace-50 & 2.941 & 1.464 & 4.173 & 6.832 & 12.247 & 23.426 \\
 & QMagFace-100 & 2.060 & 0.950 & 2.616 & 4.409 & 7.145 & 16.454 \\
 \hline
LFW & QMagFace-18 & 4.282 & 3.082 & 6.917 & 24.382 & 82.466 & 96.127 \\
 & QMagFace-50 & 2.961 & 1.572 & 4.194 & 6.815 & 12.045 & 22.677 \\
 & QMagFace-100 & 2.031 & 1.033 & 2.565 & 4.275 & 7.144 & 18.318 \\
 \hline
Morph & QMagFace-18 & 4.245 & 3.085 & 6.901 & 23.900 & 80.080 & 94.926 \\
 & QMagFace-50 & 2.939 & 1.566 & 4.141 & 6.733 & 12.130 & 24.999 \\
 & QMagFace-100 & 2.051 & 1.055 & 2.596 & 4.302 & 7.109 & 17.213 \\
 \Xhline{2\arrayrulewidth}
\end{tabular}
\end{table*}

\begin{table*}[]
\renewcommand{\arraystretch}{1.1}
\setlength{\tabcolsep}{4pt}
\centering
\caption{Evaluation on LFW based on different training datasets - The performance [$\%$] is reported in terms of FNMR at different FMRs and EER. Three MagFace variants are compared against QMagFace variants that are trained on different training sources}
\label{tab:DifferentTrainingEvalOnLFW} 
\begin{tabular}{llrrrrrr}
\Xhline{2\arrayrulewidth}
 & & & \multicolumn{5}{c}{FNMR at FMR}  \\
\cmidrule(rl){4-8}
Training database & Model &  EER  & $10^{-1}$  & $10^{-2}$  & $10^{-3}$  & $10^{-4}$  & $10^{-5}$  \\
\hline
& MagFace-18 & 1.057 & 0.324 & 1.096 & 3.710 & 9.613 & 20.163 \\
 & MagFace-50 & 0.349 & 0.110 & 0.290 & 0.462 & 0.586 & 0.821 \\
 & MagFace-100 & 0.277 & 0.159 & 0.255 & 0.297 & 0.441 & 0.621 \\
 \hline
Adience & QMagFace-18 & 0.724 & 0.186 & 0.607 & 2.324 & 7.282 & 14.377 \\
 & QMagFace-50 & 0.332 & 0.035 & 0.172 & 0.407 & 0.517 & 0.752 \\
 & QMagFace-100 & 0.195 & 0.145 & 0.172 & 0.221 & 0.331 & 0.517 \\
 \hline
ColorFeret & QMagFace-18 & 0.793 & 0.204 & 0.721 & 2.516 & 7.438 & 18.768 \\
 & QMagFace-50 & 0.304 & 0.086 & 0.199 & 0.379 & 0.516 & 0.797 \\
 & QMagFace-100 & 0.195 & 0.091 & 0.138 & 0.212 & 0.335 & 0.556 \\
 \hline
Morph & QMagFace-18 & 0.778 & 0.204 & 0.716 & 2.505 & 7.307 & 18.655 \\
 & QMagFace-50 & 0.295 & 0.086 & 0.192 & 0.357 & 0.455 & 0.662 \\
 & QMagFace-100 & 0.184 & 0.090 & 0.141 & 0.208 & 0.314 & 0.477 \\
 \Xhline{2\arrayrulewidth}
\end{tabular}
\end{table*}

\begin{table*}[]
\renewcommand{\arraystretch}{1.1}
\setlength{\tabcolsep}{4pt}
\centering
\caption{Evaluation on Morph based on different training datasets - The performance [$\%$] is reported in terms of FNMR at different FMRs and EER. Three MagFace variants are compared against QMagFace variants that are trained on different training sources}
\label{tab:DifferentTrainingEvalOnMorph} 
\begin{tabular}{llrrrrrr}
\Xhline{2\arrayrulewidth}
 & & & \multicolumn{5}{c}{FNMR at FMR}  \\
\cmidrule(rl){4-8}
Training database & Model &  EER  & $10^{-1}$  & $10^{-2}$  & $10^{-3}$  & $10^{-4}$  & $10^{-5}$  \\
\hline
 & MagFace-18 & 0.883 & 0.813 & 0.873 & 1.185 & 2.189 & 50.892 \\
 & MagFace-50 & 0.788 & 0.784 & 0.825 & 0.832 & 0.843 & 0.894 \\
 & MagFace-100 & 0.848 & 0.777 & 0.814 & 0.824 & 0.834 & 0.848 \\
 \hline
Adience & QMagFace-18 & 0.843 & 0.779 & 0.834 & 1.036 & 1.908 & 40.070 \\
 & QMagFace-50 & 0.821 & 0.473 & 0.812 & 0.826 & 0.835 & 0.880 \\
 & QMagFace-100 & 0.773 & 0.363 & 0.760 & 0.817 & 0.829 & 0.840 \\
 \hline
ColorFeret & QMagFace-18 & 0.846 & 0.790 & 0.841 & 1.059 & 1.969 & 44.750 \\
 & QMagFace-50 & 0.790 & 0.475 & 0.784 & 0.798 & 0.814 & 0.860 \\
 & QMagFace-100 & 0.763 & 0.371 & 0.747 & 0.802 & 0.813 & 0.824 \\
 \hline
 LFW & QMagFace-18 & 0.838 & 0.797 & 0.841 & 1.032 & 1.871 & 57.971 \\
 & QMagFace-50 & 0.761 & 0.477 & 0.775 & 0.795 & 0.808 & 0.845 \\
 & QMagFace-100 & 0.737 & 0.417 & 0.729 & 0.794 & 0.804 & 0.818 \\
 \Xhline{2\arrayrulewidth}
\end{tabular}
\end{table*}

\end{document}